\documentclass[conference,a4paper]{IEEEtran}
\IEEEoverridecommandlockouts

\usepackage{acro}
\usepackage{booktabs, multirow}
\usepackage{lipsum}  
\usepackage{algorithm}
\usepackage{algpseudocode}
\usepackage{placeins}
\usepackage{graphicx}
\usepackage{subcaption}
\usepackage{tabularx}
\usepackage{hyperref}
\usepackage{float}
\usepackage{booktabs, multirow}
\usepackage{cite}
\usepackage{amsmath,amssymb,amsfonts}
\usepackage{graphicx}
\usepackage{textcomp}
\usepackage{xcolor}

\def\BibTeX{{\rm B\kern-.05em{\sc i\kern-.025em b}\kern-.08em T\kern-.1667em\lower.7ex\hbox{E}\kern-.125emX}}

\DeclareAcronym{ml}{
	short = ML,
	long = {Machine Learning}
}

\DeclareAcronym{pv}{
	short = PV,
	long = {photovoltaic}
}

\DeclareAcronym{flwr}{
	short = FLWR,
	long = {A Friendly Federated Learning Framework}
}

\DeclareAcronym{mpir}{
	short = MPIR,
	long = {Mean Prediction Interval Range}
}

\DeclareAcronym{mql}{
	short = MQL,
	long = {Mean Quantile Loss}
}

\DeclareAcronym{smhi}{
	short = SMHI,
	long = {Swedish Meteorological and Hydrological Institute}
}

\DeclareAcronym{kde}{
	short = KDE,
	long = {Kernel Density Estimations}
}

\DeclareAcronym{fl}{
	short = FL,
	long = {Federated Learning}
}

\DeclareAcronym{gb}{
	short = GB,
	long = {Gradient Boosting}
}

\DeclareAcronym{lgbm}{
	short = LightGBM,
	long = {Light Gradient Boosting Machine}
}

\DeclareAcronym{qr}{
	short = QR,
	long = {Quantile Regression}
}

\DeclareAcronym{mae}{
	short = MAE,
	long = {Mean Absolute Error}
}

\DeclareAcronym{fedavg}{
	short = FedAvg,
	long = {Federated Averaging}
}

\DeclareAcronym{mape}{
	short = MAPE,
	long = {Mean Absolute Percentage Error}
}

\DeclareAcronym{cdf}{
	short = CDF,
	long = {Cumulative Distribution Function}
}

\DeclareAcronym{pi}{
	short = PI,
	long = {Prediction Interval}
}

\DeclareAcronym{iot}{
	short = IoT,
	long = {Internet of Things}
}

\DeclareAcronym{svm}{
	short = SVM,
	long = {Support Vector Machines}
}

\DeclareAcronym{ann}{
	short = ANN,
	long = {Artificial Neural Networks}
}

\DeclareAcronym{rf}{
	short = RF,
	long = {Random Forests}
}

\DeclareAcronym{cnn}{
	short = CNN,
	long = {Convolutional Neural Networks}
}

\DeclareAcronym{rnn}{
	short = RNN,
	long = {Recurrent Neural Networks}
}

\DeclareAcronym{lstm}{
	short = LSTM,
	long = {Long Short-Term Memory}
}

\DeclareAcronym{dl}{
	short = DL,
	long = {Deep Learning}
}

\DeclareAcronym{ernn}{
	short = ERNN,
	long = {Elman Recurrent Neural Network}
}

\DeclareAcronym{esn}{
	short = ESN,
	long = {Echo State Network}
}

\DeclareAcronym{sgd}{
	short = SGD,
	long = {Stochastic Gradient Descent}
}

\DeclareAcronym{flbnn}{
	short = FL-BNN,
	long = {Federated Learning-Based Bayesian Neural Network}
}

\DeclareAcronym{bnn}{
	short = BNN,
	long = {Bayesian Neural Network}
}

\DeclareAcronym{btm}{
	short = BTM,
	long = {Federated Learning-Based Bayesian Neural Network}
}

\DeclareAcronym{stann}{
	short = STANN,
	long = {Spatial and Temporal Attention-Based Neural Network}
}

\DeclareAcronym{nn}{
	short = NN,
	long = {Neural Network}
}

\DeclareAcronym{lr}{
	short = LR,
	long = {Linear Regression}
}

\DeclareAcronym{dfst}{
	short = DFST,
	long = {Distributed FastShapelet Transform}
}

\DeclareAcronym{scmfts}{
	short = SCMFTS,
	long = {Scalable and Distributed Transformation for Univariate and Multivariate Time Series}
}

\DeclareAcronym{fs}{
	short = FS,
	long = {FastShapele}
}

\DeclareAcronym{pdf}{
	short = PDF,
	long = {Probability Density Function}
}

\DeclareAcronym{ev}{
	short = EV,
	long = {Electric Vehicle}
}

\DeclareAcronym{gbdt}{
	short = GBDT,
	long = {Gradient Boosted Decision Trees}
}

\DeclareAcronym{fedprox}{
	short = FedProx,
	long = {FedProx}
}

\DeclareAcronym{feddist}{
	short = FedDist,
	long = {FedDist}
}

\DeclareAcronym{res}{
  short = RES,
  long = {renewable energy sources}
}

\DeclareAcronym{cml}{
  short = CML,
  long = {Classical Machine Learning}
}

\DeclareAcronym{dt}{
  short = DT,
  long = {Decision Tree}
}

\DeclareAcronym{cpps}{
  short = CPPS,
  long = {Cyber-Physical Systems}
}

\DeclareAcronym{picp}{
  short = PICP,
  long = {Prediction Interval Coverage Probability}
}

\begin{document}

\title{Physics-Informed State Space Models for Day-Ahead Photovoltaic Power Forecasting

\thanks{The Swedish Energy Agency has partially funded this work through the project Solar Electricity Research Centre, Sweden (SOLVE), grant number 52693-1. Additional support was provided by the Bavarian State Ministry of Science and Art through the High-Tech Agenda (HTA) and the Knowledge Foundation of Sweden (KKS) through project 20210059.

\noindent\textsuperscript{†}These authors contributed equally to this work.}
}

\author{
\IEEEauthorblockN{Phil Aupke\textsuperscript{†}\IEEEauthorrefmark{1},
Muhammad Hassan\textsuperscript{†}\IEEEauthorrefmark{2}, 
Khalid Ali\IEEEauthorrefmark{3},  
Seema\IEEEauthorrefmark{4},  
Andreas Kassler\IEEEauthorrefmark{5}\IEEEauthorrefmark{6}, and
Andreas Theocharis\IEEEauthorrefmark{8}}

\IEEEauthorblockA{
\IEEEauthorrefmark{1}\IEEEauthorrefmark{5}Department of Mathematics and Computer Science, Karlstad University, Karlstad, Sweden\\
\IEEEauthorrefmark{4}\IEEEauthorrefmark{8}Engineering and Physics Department, Karlstad University, Karlstad, Sweden\\
\IEEEauthorrefmark{2}\IEEEauthorrefmark{6}\IEEEauthorrefmark{3}Faculty of Applied Computer Science, Deggendorf Institute of Technology, Deggendorf, Germany\\
Emails: \IEEEauthorrefmark{1}phil.aupke@kau.se
\IEEEauthorrefmark{2}muhammad.hassan@stud.th-deg.de
\IEEEauthorrefmark{2}khalid.ali@th-deg.de
\IEEEauthorrefmark{4}seema.seema@kau.se\\
\IEEEauthorrefmark{5}andreas.kassler@kau.se
\IEEEauthorrefmark{6}andreas.kassler@th-deg.de
\IEEEauthorrefmark{8}andreas.theocharis@kau.se
}}

\maketitle

\begin{abstract}

Accurate day-ahead photovoltaic (PV) power forecasting requires both strong temporal modeling and adherence to physical PV system constraints. Purely data-driven models often ignore the underlying PV conversion process and may produce predictions that are outside the operational range of a PV system. In this paper, we propose a physics-informed state space framework that combines a differentiable PV conversion model with a selective state space backbone (PowerMamba). Our approach follows a physics–residual decomposition. The physics module provides a structured baseline based on solar geometry, irradiance, temperature effects, and inverter limits, while the state space model learns residual corrections from numerical weather prediction (NWP) inputs over a 24-hour horizon. Experiments on one year of utility-scale PV data show that the proposed method significantly improves forecasting accuracy, reducing RMSE by 38.5\% and MAE by 45.8\% compared to a physics-only baseline, and outperforming a purely data-driven model. Results further show that incorporating system-level physics provides the largest performance gains, while additional refinements offer only marginal improvements. Beyond accuracy, the model improves robustness by constraining predictions toward physical feasibility, reducing unrealistic predictions under corrupted measurements.

\end{abstract}


\section{Introduction}

The increasing penetration of photovoltaic (PV) generation introduces substantial uncertainty into power system operation due to its dependence on meteorological conditions. Accurate day-ahead PV forecasting is therefore essential for unit commitment, reserve allocation, and electricity market scheduling, as forecast errors directly translate into balancing costs and reduced grid reliability \cite{holderbaum2023dayahead}. Recent advances in machine learning (ML) have significantly improved PV forecasting performance by capturing nonlinear relationships between meteorological variables and power output. Deep neural networks and attention-based models have demonstrated strong predictive capabilities for renewable energy forecasting tasks \cite{zang2024adaptive, niu2025solarcast}. More recently, selective state space models (SSMs) such as Mamba have emerged as computationally efficient alternatives to Transformer architectures, enabling long-range temporal modeling with linear computational complexity \cite{gu2024mamba}. In the energy domain, PowerMamba has demonstrated strong performance across several forecasting tasks \cite{menati2024powermamba}. 

Despite their predictive accuracy, these models typically operate as black boxes and do not explicitly encode the physical constraints governing photovoltaic energy conversion. As a result, forecasts may violate system limits such as inverter capacity and can degrade under sensor drift, measurement corruption, or distribution shifts. Hybrid physics–machine learning approaches have therefore been explored for PV forecasting by combining physics-based PV performance models with data-driven correction terms or residual learning strategies \cite{yang2020hybrid, ma2019hybrid}. 

However, existing physics-informed approaches primarily incorporate limited aspects of photovoltaic behavior, such as thermal dynamics or temperature-dependent efficiency, and have not been combined with modern state space sequence architectures. In particular, the systematic integration of system-level photovoltaic conversion physics (including solar geometry, irradiance transposition, and inverter capacity constraints) with selective SSMs remains largely unexplored for operational day-ahead forecasting driven by Numerical Weather Prediction (NWP) inputs \cite{mayer2023probabilistic}.


To address this gap, we propose a physics-informed state space framework for day-ahead photovoltaic (PV) power forecasting that integrates system-level PV conversion physics with a selective state space backbone. Our approach follows a physics--residual decomposition, where a differentiable PV module captures the dominant physical processes of solar energy conversion to provide a structured baseline, while a data-driven sequence model learns residual corrections from NWP inputs over a 24-hour horizon. This design enables the model to combine physical consistency with flexible temporal modeling. Our main contributions are as follows:
\begin{itemize}
    \item We introduce a physics--residual decomposition that integrates system-level PV conversion physics with a selective SSM for NWP-driven day-ahead forecasting.
    
    \item We show through progressive ablation that core PV conversion physics delivers the dominant accuracy gain, while higher-order refinements yield diminishing returns.
    
    \item We demonstrate that the embedded physics module constrains predictions toward the inverter capacity limit, reducing unrealistic predictions under measurement corruption.
\end{itemize}

The remainder of this paper is organized as follows. Section~\ref{sec:background} reviews related work. Section~\ref{sec:methodology} presents the proposed framework. Section~\ref{sec:evaluation} reports experimental results, and Section~\ref{sec:conclusion} concludes the paper.
\section{Background and Related Work}\label{sec:background}

Recent work has explored advanced sequence modeling architectures for energy forecasting. Transformer-based approaches leverage attention mechanisms to capture long-range dependencies but suffer from quadratic computational complexity for long sequences~\cite{zang2024adaptive, niu2025solarcast}. Selective SSMs such as Mamba provide an efficient alternative with linear complexity and strong sequence modeling capability, as demonstrated by the PowerMamba architecture across multiple energy forecasting tasks~\cite{gu2024mamba, menati2024powermamba}. However, these approaches remain purely data-driven and do not encode the physical constraints governing PV energy conversion, making them susceptible to physically implausible predictions under sensor drift, measurement corruption, or distribution shifts.

Physics-informed ML has emerged as a promising paradigm for integrating domain knowledge into neural models~\cite{raissi2019physics}. In PV forecasting, several studies have incorporated simplified physical relationships such as temperature-dependent module efficiency or thermal dynamics~\cite{wang2025dynamic, han2024physics}, and hybrid frameworks combining physics-based PV performance models with ML corrections have been proposed to improve generalization~\cite{yang2020hybrid, mayer2023probabilistic}. These approaches typically incorporate only isolated physical effects and do not model the full system-level PV conversion chain, including solar geometry, irradiance transposition, and inverter capacity constraints.

In contrast, our work integrates a system-level photovoltaic conversion
pipeline with a selective state space backbone through a physics-residual decomposition for NWP-driven day-ahead forecasting.

\section{Methodology}\label{sec:methodology}

\begin{figure}[!t]
\centering
\includegraphics[width=0.9 \linewidth]{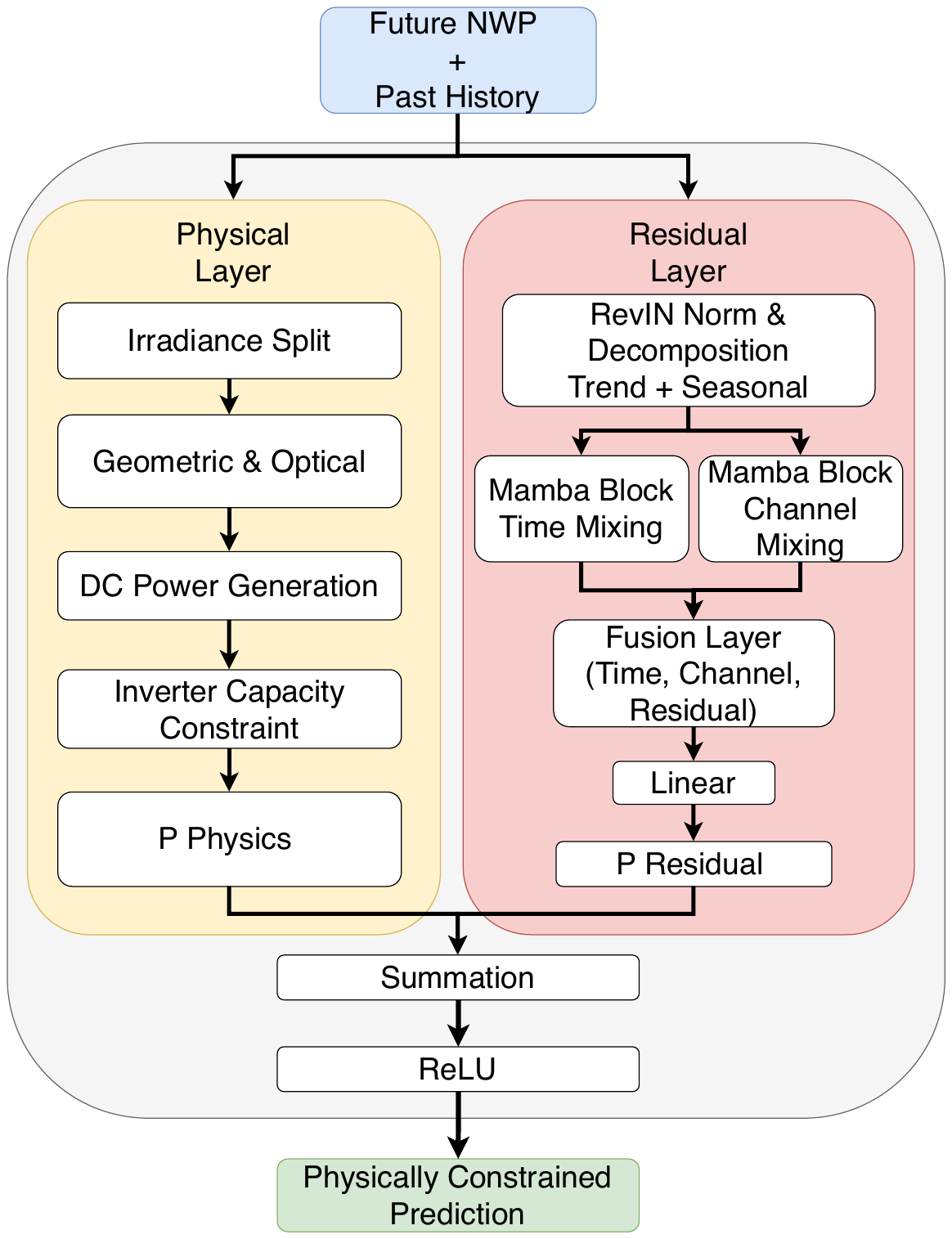}
\caption{Physics-informed PowerMamba architecture combining a differentiable PV module and a selective state space residual backbone}
\label{fig:architecture}
\end{figure}

This work addresses day-ahead PV power forecasting using a hybrid physics–residual framework that combines structured PV conversion physics with modern sequence modeling. As illustrated in Fig.~\ref{fig:architecture}, the architecture consists of three components: (i)~a \emph{Residual Layer} employing a selective state space backbone (PowerMamba) for temporal modeling of systematic deviations (Section~\ref{sec:ssm}), (ii)~a \emph{Physical Layer} implementing a differentiable PV performance module that produces a physically consistent baseline prediction through solar geometry, irradiance transposition, temperature-dependent efficiency, and inverter capacity constraints (Section~\ref{sec:physics}), and (iii)~a physics--residual summation that combines both outputs and enforces non-negative predictions via a ReLU activation (Section~\ref{sec:decomposition}).

\subsection{Problem Formulation}

The forecasting task is formulated as a multi-step sequence prediction problem. Given a historical observation window of length $L$ and forecasting horizon $H$, the model predicts PV power for the next $H$ time steps. At time index $t$, historical measurements $\mathbf{X}_{t-L+1:t} \in \mathbb{R}^{L \times D}$ and future NWP inputs $\mathbf{X}^{\mathrm{NWP}}_{t+1:t+H} \in \mathbb{R}^{H \times D_{\mathrm{NWP}}}$ produce the forecast $\hat{\mathbf{Y}}_{t+1:t+H} \in \mathbb{R}^{H}$. Table~\ref{tab:features} summarizes all input features. The historical window covers $L=288$ steps (3~days) and the prediction horizon $H=96$ steps (24~hours at 15-minute resolution). Cyclic sine--cosine encodings of hour-of-day and day-of-year capture diurnal and seasonal periodicity. Static plant parameters (latitude, longitude, tilt, azimuth, nominal capacity) are provided as configuration inputs to the physics module (Section~\ref{sec:physics}) and are not listed among the time-varying model features.

\begin{table}[!t]
\centering
\caption{Model input features}
\label{tab:features}
\renewcommand{\arraystretch}{1.1}
\setlength{\tabcolsep}{12pt}
\footnotesize
\begin{tabular}{lll}
\toprule
\textbf{Feature} & \textbf{Unit} & \textbf{Time domain} \\
\midrule
NWP global irradiance      & W/m$^2$       & Forecast horizon \\
NWP temperature            & $^\circ$C     & Forecast horizon \\
NWP wind speed             & m/s           & Forecast horizon \\
NWP wind direction         & $^\circ$      & Forecast horizon \\
NWP-derived power estimate & MW            & Forecast horizon \\
Measured PV power          & MW            & Historical window \\
Solar elevation            & $^\circ$      & Forecast horizon \\
Clear-sky power $P_\mathrm{CLR}$ & MW      & Forecast horizon \\
\bottomrule
\end{tabular}
\end{table}



\subsection{State Space Backbone}
\label{sec:ssm}

We employ a selective SSM (PowerMamba)~\cite{gu2024mamba, menati2024powermamba} to capture long-range temporal dependencies with linear complexity. Unlike Transformer-based architectures, whose quadratic attention cost becomes prohibitive for long input sequences, SSMs propagate information through input-dependent linear recurrences that scale linearly with sequence length while retaining content-dependent filtering. For a 288-step historical window and 96-step forecast horizon, this efficiency is operationally significant. The backbone evolves a latent state $\mathbf{h}_t \in \mathbb{R}^{N}$ via
\begin{align}
\mathbf{h}_{t+1} &= \mathbf{A}(\mathbf{x}_t)\,\mathbf{h}_t + \mathbf{B}(\mathbf{x}_t)\,\mathbf{x}_t, \label{eq:ssm_state}\\
\hat{y}^{\mathrm{res}}_t &= \mathbf{C}\,\mathbf{h}_t, \label{eq:ssm_out}
\end{align}
where $\mathbf{x}_t \in \mathbb{R}^{D'}$ is the input token and $\mathbf{A}(\cdot)$, $\mathbf{B}(\cdot)$ are input-dependent matrices enabling selective propagation and forgetting.

\subsection{Differentiable Photovoltaic Physics Module}\label{sec:physics}

The physics module captures the dominant deterministic processes of PV conversion while remaining fully differentiable for end-to-end training. We deliberately omit higher-order effects such as inter-row shading, soiling, spectral mismatch, and detailed inverter efficiency curves, which fall below the noise floor of 15-minute, 4\,km NWP inputs (confirmed by the ablation in Section~\ref{sec:ablation}). The key design contribution is embedding a complete, learnable PV conversion chain, from irradiance transposition through DC power generation to inverter capacity enforcement, as a differentiable layer within a neural sequence model, enabling joint optimization of physical and data-driven parameters.

Solar geometry and irradiance transposition yield plane-of-array irradiance $G_{\mathrm{POA}}(t)$. The PV module temperature, efficiency, and inverter-constrained power are then computed as:
\begin{align}
T_c(t) &= T_a(t) + \frac{G_{\mathrm{POA}}(t)}{U_0 + U_1\,v(t)}, \label{eq:faiman}\\
\eta(t) &= \eta_{\mathrm{STC}}\bigl(1 + \gamma\,(T_c(t) - 25)\bigr), \label{eq:eff}\\
P_{\mathrm{phys}}(t) &= \min\!\bigl(\eta(t)\,G_{\mathrm{POA}}(t)\,A_{\mathrm{eff}},\; P_{\mathrm{inv,max}}\bigr), \label{eq:pphys}
\end{align}
where $T_a$ is ambient temperature, $v$ wind speed, $U_0$, $U_1$ heat-transfer coefficients~\cite{faiman2008temperature}, $\eta_{\mathrm{STC}}$ module efficiency at STC, $\gamma$ the temperature coefficient, $A_{\mathrm{eff}}$ effective module area, and $P_{\mathrm{inv,max}}$ the nominal inverter rating~\cite{king2004sandia}. Thermal and efficiency parameters are learnable but bounded to physically plausible ranges, enabling site-specific adaptation while preserving consistency.

\subsection{Physics--Residual Decomposition}\label{sec:decomposition}

The decomposition separates what can be modeled from first principles (the irradiance-to-power conversion) from what must be learned from data (NWP forecast errors, site-specific effects, unresolved atmospheric variability). This separation is intended to improve generalization by preventing the
backbone from re-learning known physics, and improves robustness because
the physical baseline remains valid even when training data quality
degrades. The final prediction combines both components:

\begin{equation}
\hat{P}(t) = \mathrm{ReLU}\big(P_{\mathrm{phys}}(t) + P_{\mathrm{res}}(t)\big),
\label{eq:fusion}
\end{equation}

where $P_{\mathrm{res}}(t) = \hat{y}_t^{\mathrm{res}}$ is the state space backbone output; the ReLU enforces non-negativity, while the inverter capacity limit is enforced only within $P_{\mathrm{phys}}(t)$ via a smooth saturation (Section~III-C).

\subsection{Training Objective and Optimization}

The model is trained end-to-end using a composite loss balancing predictive accuracy and physical consistency:
\begin{equation}
\mathcal{L} = \mathcal{L}_{\mathrm{MSE}} + \lambda_{\mathrm{phys}}\,\mathcal{L}_{\mathrm{phys}},
\end{equation}

where $\mathcal{L}_{\mathrm{MSE}}$ is the mean squared error between predicted and measured PV power and $\mathcal{L}_{\mathrm{phys}} = \sum_{k \in \Theta} \max(0,\, \theta_k - \theta_k^{\max}) + \max(0,\, \theta_k^{\min} - \theta_k)$ penalizes learned physics parameters $\Theta = \{\eta_{\mathrm{STC}}, \gamma, U_0, U_1\}$ leaving physically plausible bounds ($\lambda_{\mathrm{phys}} = 0.01$).

Training follows a two-stage strategy over 30 epochs (batch size 64, Adam, lr $10^{-3}$): during the first 10 warm-up epochs the residual backbone is frozen while the physics module is pre-fitted to the irradiance-to-power relationship; the backbone then activates for joint training, preventing the data-driven component from dominating the physical baseline. The backbone employs two Mamba blocks ($d_{\mathrm{model}}=128$, $d_{\mathrm{state}}=64$, $d_{\mathrm{conv}}=2$, expansion factor 2, dropout $p=0.2$). Input features are normalized via Reversible Instance Normalization (RevIN) within the model.
\section{Evaluation}\label{sec:evaluation}

The proposed framework is evaluated on day-ahead photovoltaic power forecasting using real-world operational data. Experiments assess three aspects: overall forecasting accuracy, the impact of progressively embedded physical components, and robustness under measurement corruption.

\subsection{Experimental Setup}

The framework is evaluated on one year of utility-scale photovoltaic power data from a PV plant in Hebei Province, China (July 2018 -- June 2019), with an effective inverter capacity limit of approximately 13.2~MW. The dataset spans a full annual cycle, covering high-irradiance summer, low-irradiance winter, and transitional seasons, enabling evaluation under diverse meteorological conditions. Power output is recorded at 15-minute resolution together with ground-based meteorological measurements and NWP forecasts at approximately 4~km spatial resolution, yielding a 24-hour forecasting horizon (96 steps).

Performance is evaluated using rolling time-series cross-validation with four chronological folds using an expanding training window: Fold~1 trains on 20\% of the data, Fold~2 on 40\%, Fold~3 on 60\%, and Fold~4 on 80\%, with each test set covering approximately one season. Root mean squared error (RMSE) and normalized RMSE (nRMSE) are computed over daylight hours (solar elevation $> 0°$) to prevent bias from nighttime zero-production.

\begin{table}[!b]
\centering
\caption{Day-ahead forecasting performance across all validation folds}
\label{tab:accuracy}
\renewcommand{\arraystretch}{1.2}
\setlength{\tabcolsep}{2pt}
\footnotesize
\begin{tabular}{lccc}
\toprule
\textbf{Model} & \textbf{RMSE [MW]} & \textbf{MAE [MW]} & \textbf{nRMSE [\%]} \\
\midrule
Physics-only (NWP-driven)      & 2.651 & 1.396 & 13.21 \\
Smart persistence              & 2.272 & 1.167 & 11.36 \\
Data-driven PowerMamba         & 2.035 & 0.996 & 10.15 \\
Physics-informed PowerMamba    & \textbf{1.631} & \textbf{0.756} & \textbf{8.16} \\
\bottomrule
\end{tabular}
\end{table}

\begin{figure}[t]
\centering
\includegraphics[width=\linewidth]{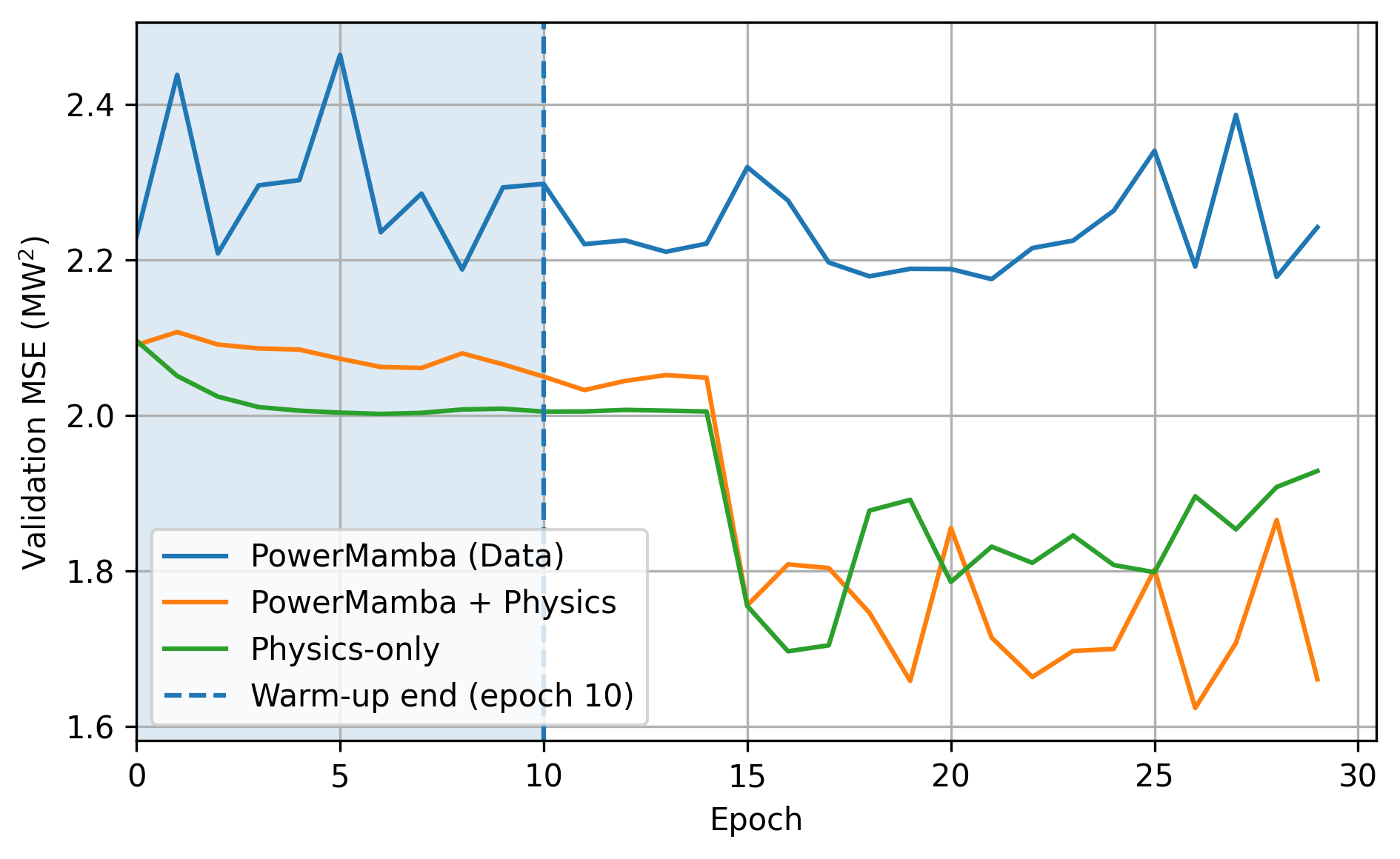}
\caption{Validation loss curves for Fold~2 comparing the physics-only, data-driven, and physics-informed models}
\label{fig:convergence}
\end{figure}

\subsection{Physics-Informed State Space Architecture}

We compare the proposed model against three baselines: (i) a deterministic physics-only PV performance model driven by NWP inputs, (ii) a smart persistence model that assumes constant atmospheric transmittance while accounting for solar geometry, and (iii) an identical selective state space model (PowerMamba) without embedded physics. The proposed physics-informed PowerMamba significantly outperforms all baselines (Table~\ref{tab:accuracy}), achieving a 38.5\% RMSE reduction and 45.8\% MAE reduction relative to the physics-only model, and a 28.2\% RMSE reduction relative to smart persistence. Paired t-tests across the four folds are indicative of a difference
(p = 0.030 vs.\ physics-only, p = 0.004 vs.\ data-driven), given the small
number of folds evaluated. The large gap to the physics-only baseline indicates that the NWP-driven conversion chain alone cannot capture site-specific losses, transient cloud effects, and systematic NWP biases, which the residual backbone learns effectively. Fold-wise evaluation confirms stable generalization across seasons ($\sigma_{\mathrm{RMSE}} = \pm 0.02$~MW for Folds~1--3), with the elevated Fold~4 error attributable to measurement corruption (Section~\ref{sec:robustness}). Figure~\ref{fig:convergence} reports validation MSE (MW$^2$) for the three models on Fold 2. The physics-informed curve tracks the physics-only curve during the
epoch 0--10 warm-up (Section III-E), then drops sharply once the
residual backbone activates, stabilizing below both baselines; the
physics-only curve's larger oscillation reflects its lack of a
data-driven correction term.

\subsection{Assessment of Embedded Physical Components}\label{sec:ablation}

To quantify the impact of individual physical components, progressive ablation experiments were conducted (Table~\ref{tab:ablation}). The results reveal a clear hierarchy: solar geometry alone yields a
7.1\% RMSE reduction, temperature-dependent and system-level effects
bring the cumulative reduction to 13.3\%, and atmospheric refinements
add a further 0.6\% on top. This diminishing return is expected: at 15-minute temporal and 4 km NWP spatial resolution, higher-order atmospheric effects such as aerosol optical depth and spectral corrections are poorly resolved in the input data, limiting what the residual
branch can learn from them at this resolution.

\begin{table}[!t]
\centering
\caption{Impact of progressively embedded physical components}
\label{tab:ablation}
\renewcommand{\arraystretch}{1.2}
\footnotesize
\begin{tabular}{lcc}
\toprule
\textbf{Model Variant} & \textbf{RMSE [MW]} & \textbf{Improvement [\%]} \\
\midrule
Data-driven baseline                & 2.035 & -- \\
+ Solar geometry                    & 1.890 & 7.1 \\
+ Thermal effects (system-level)    & 1.764 & 13.3 \\
+ Atmospheric refinements           & 1.753 & 0.6 \\
\bottomrule
\end{tabular}
\end{table}

\begin{figure*}[t]
\centering
\includegraphics[width=0.8\linewidth]{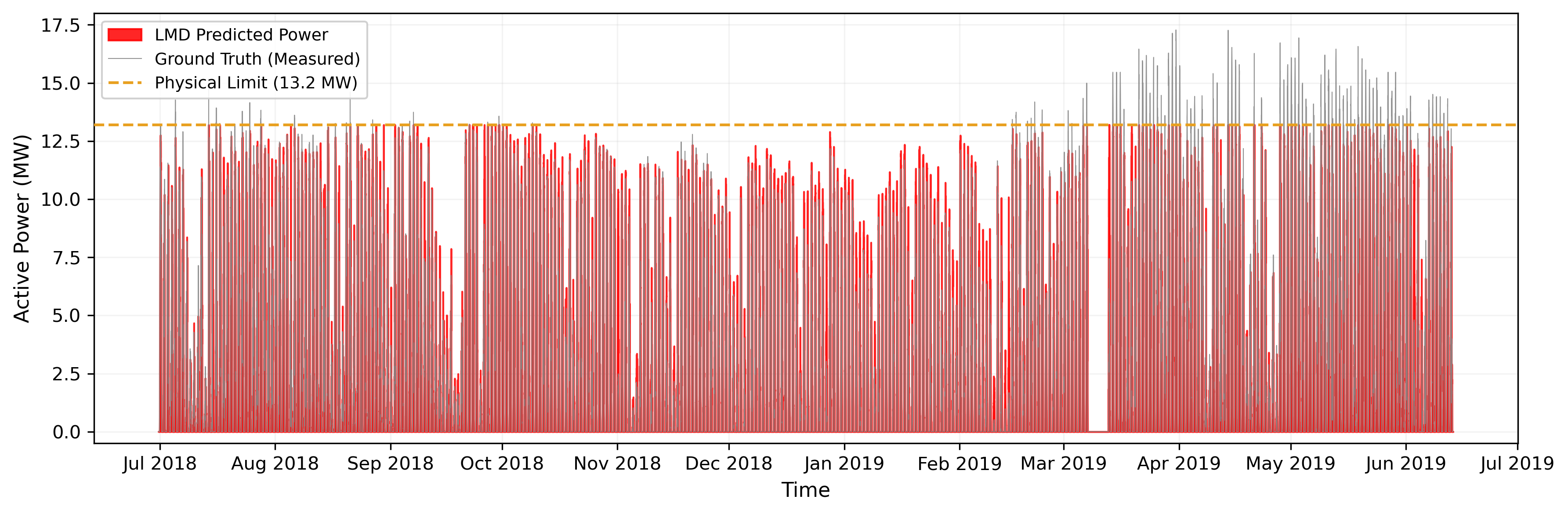}
\caption{Temporal evolution of measured power (grey) and physics-based predictions from NWP inputs (red). The dashed line marks the inverter capacity limit (13.2~MW). After March 2019 (Fold 4), measured peaks exceed this limit, while the
physics-informed model's predictions remain within the inverter constraint, avoiding unrealistic spikes}
\label{fig:corruption}
\end{figure*}

\subsection{Robustness and Physical Consistency}\label{sec:robustness}

Unlike purely data-driven models, the proposed framework constrains
predictions toward physical feasibility even under corrupted training
labels. The dataset exhibits an apparent measurement corruption period after
March 2019 during which recorded power peaks reach 17.28 MW, exceeding the physical inverter
capacity limit of 13.2 MW by 30.9\%, as shown in Fig. 3. This also
exceeds the plant's realistic operating ceiling ($\sim$15.8 MW, i.e. 79\%
of the 20 MW nameplate capacity) by 9.4\%, consistent with a sensor or
SCADA scaling error. As $P_{\text{inv,max}}$ is calibrated once and held fixed, the model
cannot distinguish this from a genuine capacity change without
periodic recalibration on updated operating data. A small number of isolated above-limit readings also occur before March 2019 (median 13.5 MW, $<$0.3\% of pre-corruption samples),
consistent with minor sensor noise rather than systematic error.

The primary performance results reported in Tables~\ref{tab:accuracy}--\ref{tab:ablation} are therefore based predominantly on clean training data: Folds~1--3 train exclusively on pre-corruption periods, and only Fold~4 incorporates systematically corrupted measurements in its training set. A purely data-driven PowerMamba baseline, trained on corrupted targets
in Fold 4, learns to predict physically implausible power levels above the inverter limit. In contrast, the physics module caps predictions at approximately 12.5 MW regardless of measurement quality. This effective cap, slightly below the 13.2 MW nameplate inverter rating, emerges from
the learned physics parameters and reflects the plant's realistic
operating ceiling after conversion losses. Across Fold 4, the
data-driven PowerMamba baseline produces above-limit predictions in
26.9\% of daylight steps, whereas the physics-informed model's
predictions remain within the inverter capacity constraint. The elevated Fold~4 RMSE of 2.02~MW compared to $1.60 \pm 0.02$~MW for Folds~1--3 therefore reflects evaluation against inflated ground-truth targets rather than model degradation. Excluding Fold~4, the physics-informed model achieves a mean RMSE of 1.50~MW (7.50\% nRMSE). This demonstrates robustness to real-world data quality problems that are common in operational PV monitoring systems.
\section{Conclusion and Future Work}\label{sec:conclusion}

Embedding system-level physics is the key factor for improving day-ahead PV forecasting accuracy and robustness. This paper presented a physics-informed state space framework for day-ahead photovoltaic power forecasting that integrates a differentiable PV conversion module with a selective state space backbone (PowerMamba) through a physics--residual decomposition. Experiments on utility-scale PV data demonstrate a 38.5\% RMSE and 45.8\% MAE reduction over a physics-only benchmark, with ablation results confirming that system-level physics provides the dominant gain while atmospheric refinements offer only marginal improvements. The embedded physics further improves robustness by preventing physically implausible predictions during measurement corruption. The physics module adds only 141 parameters to the 517k-parameter backbone with no added sequential depth; combined with the backbone's linear-time complexity, full 24-hour inference completes in seconds on a single GPU. Only the physics parameters would likely require recalibration for
new sites, though this has not been verified beyond the single site
studied here.

As these results are derived from a single utility-scale installation, the magnitude of the reported improvements and the relative importance of individual physics components may not generalize to sites with different climates, array configurations, or inverter architectures, and variant comparisons were not verified with repeated-seed runs.

\bibliographystyle{IEEEtran}
\bibliography{refs}

\end{document}